 \documentclass[final,3p,times,authoryear]{elsarticle}
\usepackage{amssymb}
\journal{}
\usepackage{booktabs}
\usepackage{float}
\usepackage{multirow}
\usepackage{array}
\usepackage{hyperref}
\hypersetup{colorlinks=true}
\usepackage[ruled,linesnumbered]{algorithm2e}

\SetAlgorithmName{Algorithm}{Algorithm}{Algorithm}

\begin{document}

\begin{frontmatter}

\title{A concise method for feature selection via normalized frequencies}

\author[1]{Tan Song}
\author[2]{Xia He}

\address[1]{Xihua University}
\address[2]{Xihua University}

\begin{abstract}
Feature selection is an important part of building a machine learning model. By eliminating redundant or misleading features from data, the machine learning model can achieve better performance while reducing the demand on com-puting resources. Metaheuristic algorithms are mostly used to implement feature selection such as swarm intelligence algorithms and evolutionary algorithms. However, they suffer from the disadvantage of relative complexity and slowness. In this paper, a concise method is proposed for universal feature selection. The proposed method uses a fusion of the filter method and the wrapper method, rather than a combination of them. In the method, one-hoting encoding is used to preprocess the dataset, and random forest is utilized as the classifier. The proposed method uses normalized frequencies to assign a value to each feature, which will be used to find the optimal feature subset. Furthermore, we propose a novel approach to exploit the outputs of mutual information, which allows for a better starting point for the experiments. Two real-world dataset in the field of intrusion detection were used to evaluate the proposed method. The evaluation results show that the proposed method outperformed several state-of-the-art related works in terms of accuracy, precision, recall, F-score and AUC. 
\end{abstract}

%\begin{highlights}
%\item 	A concise method for feature selection is proposed.
%\item 	A novel approach is proposed to exploit the outcomes of the filter method.
%\item 	The proposed method does not require the use of iteration.
%\item 	The proposed method outperformed several state-of-the-art feature selection methods.
%\item 	The principles of the proposed method are illustrated using experiments.
%\end{highlights}


\end{frontmatter}

\section{Introduction}
With the gradual expansion of feature space, it is more and more difficult for people to recognize feature space \citep{R1}. The irrelevant or misleading features will degrade the performance of the classifier. The emergence of feature selection has turned these problems around. To solve these problems, feature selection is used to improve the performance of the classifier by removing the irrelevant and misleading features from original features. \citep{guyon2003an}.
	
Feature selection is helpful to reduce the dimension of feature space, which can reduce the computational cost, improve the performance of classifier and restrain the occurrence of over fitting. The design of feature selection methods is to select better features that allow the classifier to achieve better performance while reducing the demand on computing resources \citep{R3}. 

The filter method and the wrapper method are two main  feature selection motheds \citep{R4}. The former ranks features and selects the best part of features as final feature subset. Mathematical methods are mostly used to measure the relationship between each feature and label. An evaluation value is calculated to each feature, which is used to rank the features \citep{R5}. The wrapper method sorting the feature subsets, the best one is as the final feature subset. The feature subsets are generated by the method and an evaluation value obtained by the classifier  is used to rank the feature subsets \citep{R6}.
	
The metaheuristic algorithms are proposed for the optimization problems \citep{R7}. A deterministic algorithm can obtain the optimal solution to the optimization problem, while a metaheuristic algorithm is based on an intuitive or empirical construction that can give a feasible solution at an acceptable cost, and the degree of deviation of that feasible solution from the optimal solution may not be predictable in advance \citep{R8}.
	
The feature selection is essentially a Non-deterministic Polynomial (NP) problem, which is solved by the metaheuristic algorithms \citep{Yusta2009Different}. Metaheuristic algorithms rely on a combination of local and global search to find an optimal solution in a large solution space. The search process requires the use of iteration to approach the optimal solution, and the setting of the parameters in search process has a significant impact. Both advanced algorithms and suitable parameters are needed to achieve a favorable solution.

As mentioned above, a sophisticated design is necessary for the metaheuristic algorithm to balance local and global search. This design trades relative complexity for the validity of the algorithm, and different tasks require individual finding of the suitable parameters. Swarm intelligence algorithms are gradually becoming the main implementation of metaheuristic algorithms such as \citep{R9,Yusta2009Different,R91}.
	
The main contributions of this paper are summarized as follows:
\begin{enumerate}[1.]

\item Use one-hot encoding to process categorical features and perform feature selection directly from the processed high-dimensional feature space.
\item Propose a novel approach to exploit the outcomes of filter method.
\item Propose a concise method for feature selection.

\end{enumerate}

The rest of this paper is organized as follows: \autoref{S2} reviews the related works. \autoref{S3} introduces two powerful tools used in this paper. \autoref{S4} details the proposed feature selection method. \autoref{S5} discusses the experiments and results. \autoref{S6} concludes this paper.

\section{Related works}
\label{S2}
The features in the dataset are not independent. In the context,it is essential for feature selection to consider the interaction between features. \cite{R6} explored the interaction between features. In this paper, authors investigated the strengths and weaknesses of the wrapper method and provided some improved design solutions. The wrapper method is designed to find the optimal feature subset.During the experimen, performance evaluation is based on some datasets. The experimental results indicate that the proposed algorithm achieves an improvement in accuracy.

With feature selection going from the edge of the stage to the center, \cite{guyon2003an}	provided an introduction of feature selection. In this paper,the following aspects of knowledge are discussed: the definition of the objective function,feature construction, feature ranking, multivariate feature selection, feature validity evaluation method and efficient search methods for better feature subset. Datasets in many areas have thousands of features, which makes feature selection especially useful for two purposes: better the performance of classifier, faster the speed of classifier.

\cite{R10} developed a noval feature selection method to overcome the limitations of MIFS \citep{R11}, MIFS-U \citep{R12},and mRMR \citep{R13}. The model known as NMIFS (normalized mutual information feature selection) is designed to optimize the measure of the relationship between features and labels. NMIFS is a filter method independent of any machine learning model. The purpose of normalization is to reduce the bias of mutual information toward multivalued features and restrict its value to the interval [0,1]. NMIFS does not require user-defined parameters such as $\beta$ in MIFS and MIFS-U. Compared to the MIFS, MIFS-U, and mRMR. NMIFS perform better on multiple artificial datasets and benchmark problems. In addition, the authors combine NMIFS with genetic algorithm and propose the GAMIFS, which uses NMIFS to initialize a better starting point and as part of a mutation operator. During the mutation process,features with high mutual information value will have a higher probability of being selected, which speeds up the convergence of the genetic algorithm.

The rough sets for feature selection were proven to be feasible \citep{R16}. The particle swarm optimization algorithm is an excellent metaheuristic algorithm \citep{R17}. \cite{R18} proposed an algorithm based on rough sets and particle swarms. The particle swarm optimization algorithm uses a number of particle flights in the feature space by interparticle interactions to find the best feature subset. The proposed algorithm utilized UCI datasets \citep{R19} for evaluation. The experimental results are compared with a GA-based approach and other deterministic rough set reduction algorithms. The experimental results showed that the proposed algorithm produces better performance.

Intrusion detection system (IDS) is an important security device. \cite{R20} proposed a mutual information-based feature selection method for IDS. A robust intrusion detection system needs to be both high performance and high speed. The proposed method uses LSSVM for classifier construction. The KDD Cup 99 dataset is used for the evaluation process and the evaluation results indicate that the proposed method produces a high level of accuracy, especially for remote to login (R2L) and user to remote (U2R) attacks.

Salp swarm algorithm (SSA) is an excellent optimization algorithm designed for continuous problems \citep{R21}. \cite{R22} came up with a novel approach which combines SAA with chaos theory (CSAA). Simulation results showed that chaos theory  improves the convergence speed of the algorithm significantly. The experiments reveal the potential of CSAA for feature selection, which can select fewer features while achieving higher classification accuracy. 

Water wave optimization(WWO) is new nature-inspired metaheuristic algorithm that was developed by \citep{R23}.The approach known as WWO simulates the phenomena of water waves refraction, propagation and fragmentation to find the global optimal solution in optimization problems. A new feature selection algorithm uses a combination of rough set theory (RST) and  a binary version of the water wave optimization approach (WWO) was proposed by \citep{R24}. Several datasets are used to evaluate the proposed algorithm, and the results is compared to several advanced metaheuristic algorithms. The computational results demonstrate the efficiency of the proposed algorithm in feature selection.

Finally,the PIO (Pigeon Inspired Optimizer) is an advanced bionic algorithm proposed by \citep{R25}, \cite{R26} proposed two binary schemes to improve PIO to accommodate the feature selection problem and applied it to intrusion detection system. Three popular datasets: KDD Cup 99, NSL-KDD, and UNSW-NB15, are used to test the algorithm. The proposed algorithm surpasses six state-of-the-art feature selection algorithms in terms of F-score and other metrics. Further, Cosine\_PIO selecte 7 features from KDD Cup 99, 5 features from NSL-KDD and 5 features from UNSW-NB15. It is amazing to achieve a excellent performance by so few features.

In the field of feature selection, classifiers aside, the wrapper methods are gaining popularity with the development of computer hardware. Various metaheuristic and hybrid feature selection methods have been proposed. However, these algorithms have some shortcomings: hard to understand and learn; more difficult to determine parameters; large computational cost.

To solve the above problems, the paper proposed a concise method for feature selection via normalized frequencies (\textbf{NFFS}). This method can perform feature selection at a much lower computational cost while maintaining high performance. The best feature is that it has a very simple logic, so it can be applied easily.

\section{One-hot encoding and mutual information}
\label{S3}
	This section presents two tools used in this paper. Processing categorical features is an indispensable preprocessing step in machine learning, which is performed by One-hot encoding in this paper. Mutual information is an excellent filter method that can capture both the linear and nonlinear dependencies between feature and label.
\subsection{One-hot encoding}
One-hot encoding encode categorical features as a one-hot numeric array \citep{R241}. One-hot encoding projects the categorical features to a high-dimensional feature space. It allows the distance among features to be calculated more reasonably, which is important for many classifiers. 

\begin{figure}
	\includegraphics[width=1\textwidth]{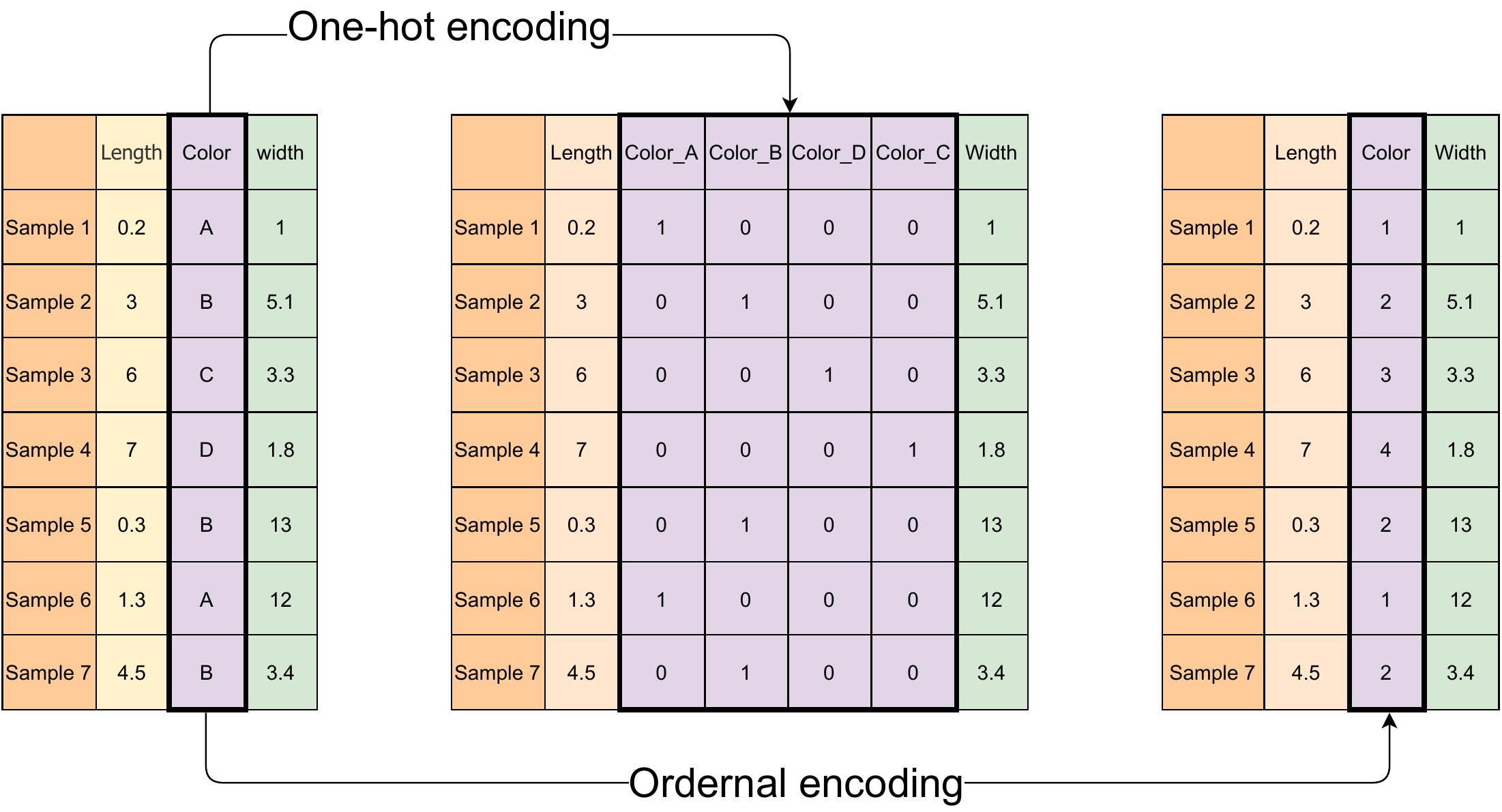}
	\caption{Comparison of one-hot encoding and ordinal encoding.}
	\label{F1}
\end{figure}

One-hot encoding uses $N$ status registers to encode $N$ states. Each register has its own individual register bits and only one of which is valid at any given time. It's easier to understand one-hot encoding with an example, let's side there is a dataset of a household item with seven samples. Length, Width and Color are used to describe each sample, but Color is not a numerical feature so it needs to be transformed. Ordinal encoding is a common encoding approach for categorical features \citep{R19}, by which feature is converted to ordinal integers. The contrast between one-hot encoding and ordinal encoding is illustrated in \autoref{F1}.

As \autoref{F1} shows, one-hot encoding transforms the feature ‘Color’ into four features ( Color\_A, Color\_B, Color\_C, Color\_D). The characters ( Non-numeric ) A, B, C and D indicate different colors. From the middle table in  \autoref{F1}, it can be noted that each sample takes 1(numeric) in one of the four color features, while 0 is taken in the other three features ‘Color\_’. As can be seen from \autoref{F1}, one-hot encoding will expands the dimensionality of the dataset  in comparison to ordinal coding. We also use \_ as a separator in the processing of the dataset in the later part of this paper.

\subsection{mutual information}

Mutual information can be applied for evaluating dependency between random variables \citep{R251}. Let $X$ ( feature ) and $Y$ ( label ) be two discrete random variables, The mutual information (\textbf{MI} value) between $X$ and $Y$ can be calculated by \autoref{E1}.
\begin{equation}
\label{E1}
\large
I(X;Y)=\sum_{x\in X}^{}\;\sum_{y\in Y}^{}p(x,y)\log\frac{p(x,y)}{p(x)p(y)}
\end{equation}

Where $I(X; Y)$ is mutual information, $p(x,y)$ is the joint probability density function, $p(x)$ and $p(y)$ are marginal density functions of $X$ and $Y$, respectively.  From the equation, we know that when $X$ and $Y$ are independent of each other, their MI value is 0, otherwise it must be greater than 0.

\section{ NFFS}
\label{S4}
This section details the proposed feature selection method in two phases. Phase I of NFFS is described in \autoref{S4.1} and phase II of NFFS is located in \autoref{S4.2}. Phase II further processes the information provided by phase I and finally finds the best feature subset. \autoref{T1} lists some abbreviations appeared in this paper, which allow the paper more concise and clear.

NFFS is different from common feature selection methods in the following two points:
\begin{enumerate}
\item NFFS selects features from the feature space of the preprocessed dataset, rather than from the raw feature space. The features selected by NFFS will not contain any categorical features.
\item All steps of NFFS use the preprocessed dataset. Only the preprocessing process would touch the raw dataset. The dimensionality of the preprocessed dataset will be greater than the raw dataset.
\end{enumerate}

\begin{table}
\centering
\caption{Abbreviations used in this paper and their meanings.}
\label{T1}
\begin {tabular}{cccc} 
\toprule
WV1 & AFS1 & WV2  & AFS2 \\ \midrule
\begin {tabular} [c]{@{}c@{}}Weight vector\\ obtained \\ in phase I of NFFS\end {tabular} & \begin {tabular}[c]{@{}c@{}}Alternative feature subsets\\ generated\\ in phase I of NFFS\end {tabular} & \begin {tabular}[c]{@{}c@{}}Weight vector\\ obtained \\ in phase II of NFFS\end {tabular} & \begin {tabular}[c]{@{}c@{}}Alternative feature subsets\\ generated\\ in phase II of NFFS\end {tabular} \\ \bottomrule 
\end{tabular}
\end{table}

\begin{figure}[h]
	\centering
	\includegraphics[width=\textwidth]{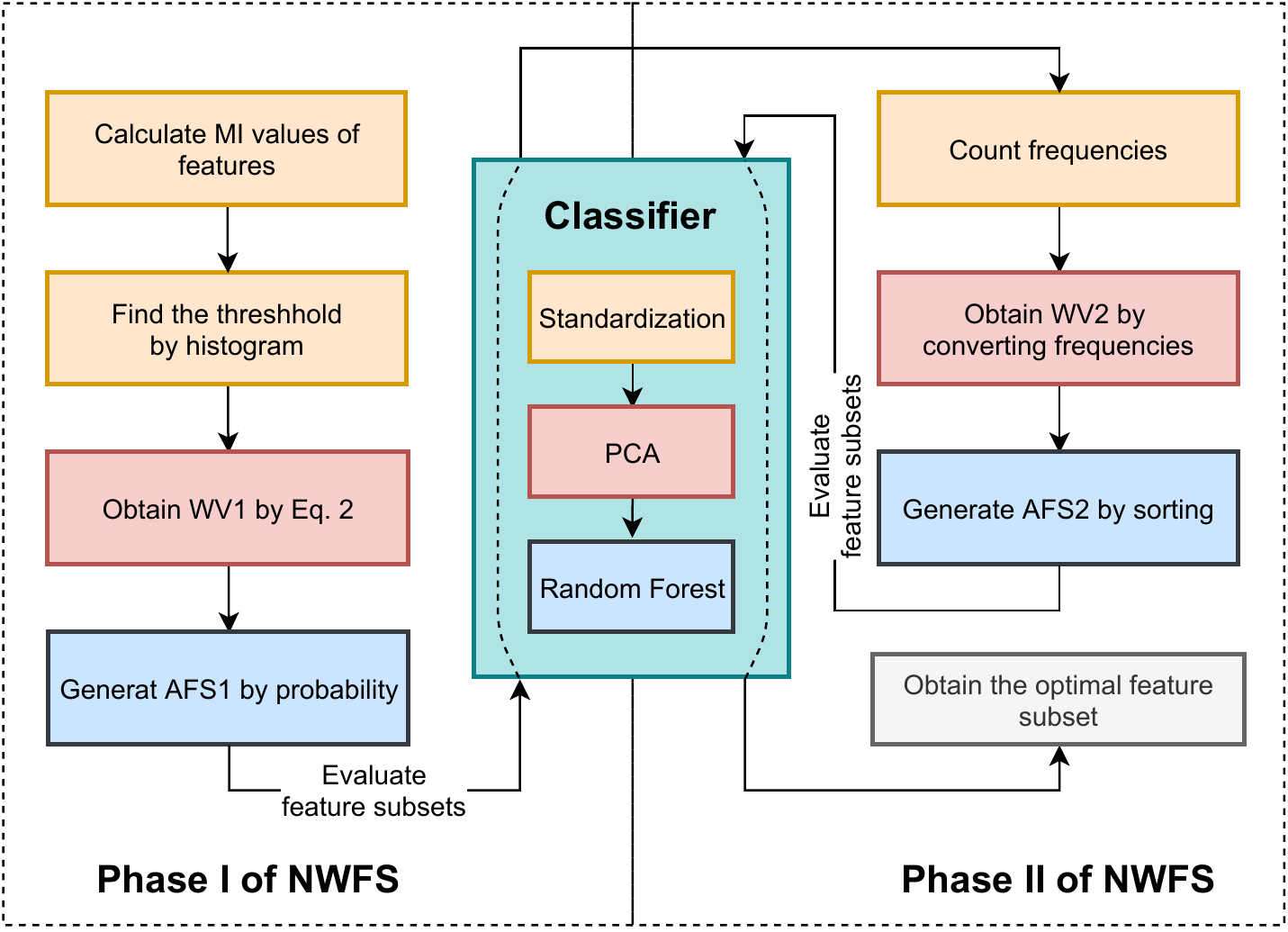}
	\caption{NFFS for feature selection.}
	\label{F2}
\end{figure}

\subsection{Phase I of NFFS}
\label{S4.1}
	The purpose of phase I of NFFS is to generate a batch of feature subsets, a portion of which can yield a slightly higher fitness values. The mechanism of this section is shown at the left side of \autoref{F2}, the steps being as follows:

\textbf{1. measure MI values}: Mutual information is used here only to evaluate the MI value for each feature. The preprocessed dataset is fed to the mutual information module, which outputs a positive floating number for each feature. The larger the number is, the stronger the relationship between feature and label, and vice versa.

\textbf{2. Find the threshold}: In the above step we obtained the MI value for each feature in the dataset. Histogram is used to analyze the distribution of MI values of features in the dataset. The distribution for these MI values of NSL-KDD dataset (A dataset used for the experiment) is shown in \autoref{F4}, from which it can be noticed that the majority of features obtained tiny MI values. A threshold is introduced to filter out these tiny MI values since it is not necessary to be calculated in the next step.

Finding the threshold requires analyzing MI values of features first, and the histogram is a handy tool. It is important to note that the threshold is selected by analyzing the histogram rather than a self-defined parameter.

\textbf{3. Obtain WV1}: A formula is used to convert MI values of features into weights of features, which is defined as in \autoref{E2}.
\begin{equation}
\label{E2}
\large
\mathrm{WV1}_i=\left\{\begin{array}{lc}\left(V_i-V_t\right)\frac{0.4}{V_{max}-V_t}+0.5&V_i>V_t\\0.5&V_i\leq V_t\end{array}\right.
\end{equation}

Where $i$ (number of features) denotes the $i$-th item of a vector, vector WV1 (weight vector obtained in phase I of NFFS) denotes weights of features, $V_i$ denotes MI value of the $i$-th feature. $V_{max}$ denotes the maximum MI value in vector $V$, while $V_t$ is the threshold from the previous step. Where 0.4 represents the upper bound for the increase of weights, which is intended to keep the weights in a suitable range for next step. Another constant, 0.5, denotes the basic weight that all features can hold.

It is clear from \autoref{E2} and \autoref{F4} that there are a large number of features with a weight of 0.5, but this is not a problem, phase I of NFFS isn't about getting an awesome result.

\textbf{4. Generate AFS1 by \emph{probability}}: When the WV1 is obtained, it's also the time to generate feature subsets. The generation process requires the use of \autoref{E3}.

\begin{equation}
\label{E3}
\large
\mathrm{AFS1}_L^i=sgn(\mathrm{WV1}_i-rand_i)=\left\{\begin{array}{lr}1&\mathrm{WV1}_i-rand_i>0\\0&\mathrm{WV1}_i-rand_i\leq0\end{array}\right.
\end{equation}

Where $L$ denotes the $L$-th generated feature subset, $i$ denotes the $i$-th item of a vector. $AFS1_L^i$ denotes the $i$-th feature of the $L$-th feature subset. AFS1$_L$ , WV1 and rand are all vector with the same dimension as the number of features in th dataset. AFS1$_L$ is a mask to represent a features subset. Each item in $rand$ is a uniform random number in the range [0,1]. In the vector AFS1$_L$, `1' means the feature is selected, while `0' means the feature is not selected.

Applying \autoref{E3}, the $L$ feature subsets constitute AFS1 (alternative feature subsets generated in phase I of NFFS). From \autoref{E3}, it can be learned that features with higher MI value will have a higher probability of being selected,  as a result, the introduction of mutual information makes NFFS have a better starting point. 

\textbf{5. Evaluate feature subsets}: Each feature subset in AFS1 is evaluated using classifier to obtain a fitness value. Specifically, the evaluation process consists of the following three steps: first, prepare training dataset and testing dataset according to AFS1$_L$; next, train the classifier with the training set; finally, test dataset is used to evaluate the trained classifier, and the result is the fitness value of AFS1$_L$.

As \autoref{E2} demonstrates, we use the outcomes of mutual information directly, rather than as a filter method to rank features. Phase I of NFFS is now complete, feature subsets in AFS1 and their fitness values are the raw materials for phase II of NFFS. 

\autoref{Phase I of NFFS} show an overall procedure for Phase I of NFFS. As the algorithm shows that there are two layer while loops. The first layer represents a generated feature subset, while the second layer determines which features are selected in this feature subset.

Also we list the shapes of some of the variables from \autoref{Phase I of NFFS} in \autoref{Variables}, taking the NSL-KDD dataset as an example. \autoref{Variables} also contains some of the variables in \autoref{Phase II of NFFS}.

\begin{algorithm}
\caption{Phase I of NFFS.}
\label{Phase I of NFFS}
\KwIn{$L$}
\KwResult{AFS1; fitness values of AFS1}
Use mutual information to score each feature.\\
Select an appropriate threshold value.\\
Use \autoref{E2} to get WV1.\\
\While{$L>0$}{
\While{$i>0$}{Get $AFS1_L^i$ by \autoref{E3}.\\$i=i-1$}
$L=L-1$}
Evaluate the feature subsets inside AFS1.\\
\Return AFS1; fitness values of AFS1
\end{algorithm}

\begin{table}[]
\centering
\caption{Shapes of partial ariables in \autoref{Phase I of NFFS} and \autoref{Phase II of NFFS} }\label{Variables}
\begin{tabular}{ccccccc}
\toprule
\textbf{value} & \textit{WV1} & \textit{$WV2$} & \textit{AFS1} & \textit{$AFS2$} & \textit{\begin{tabular}[c]{@{}c@{}}fitless values\\ of AFS1\end{tabular}} & \textit{\begin{tabular}[c]{@{}c@{}}fitless values\\   of $AFS2$\end{tabular}} \\ \midrule
\textbf{shape} & 1x122        & 1x122        & $L$x122       & $O$x122        & $L$x1                                                                     & $O$x1                                                                        \\ \bottomrule
\end{tabular}
\end{table}

\subsection{Phase II of NFFS}
\label{S4.2}
A new weight vector (WV2) would be obtained by utilizing the raw materials provided by phase I of NFFS, which is the secret sauce for finding the optimal feature subset. The mechanism of this section is shown at the right side of \autoref{F2}, the steps being as follows:

\textbf{1. Count frequencies}: First, sort the feature subsets in AFS1 by their fitness values, the $M$ feature subsets with higher fitness values constitute AFS1$_{top}$, the $N$ $(M+N<L)$ feature subsets with lower fitness values constitute AFS1$_{bottom}$, the rest feature subsets with ordinary fitness values are not involved in the counting. Next, count how many times each feature be selected in AFS1$_{top}$ and AFS1$_{bottom}$, respectively. The counting results constitute vector $\overrightarrow{F_{top}}$ and vector $\overrightarrow{F_{bottom}}$, respectively. The dimensions of these two vectors are the same as the number of features in dataset. The i-th item in $\overrightarrow{F_{top}}$ represents the total number of occurrences of the i-th feature of the dataset in AFS1$_{top}$. The i-th item in $\overrightarrow{F_{bottom}}$ represents the total number of occurrences of the i-th feature of the dataset in AFS1$_{bottom}$. 

\textbf{2. Obtain WV2}: WV2 can be derived from $\overrightarrow{F_{top}}$ and $\overrightarrow{F_{bottom}}$ by \autoref{E4}.

\begin{equation}
\label{E4}
\large
\mathrm{WV2}=\frac{\overrightarrow{F_{top}}}{\left\|\overrightarrow{F_{top}}\right\|}-\frac{\overrightarrow{F_{bottom}}}{\left\|\overrightarrow{F_{bottom}}\right\|}
\end{equation}

Where $\left\|\overrightarrow{F_{...}}\right\|$ denotes the length of a vector, the purpose of which is to normalize the vector in order to obtain a normalised frequency. The normalized frequencies have the same effect as the weights. The logic of the equation is that if a feature appears often in AFS1$_{top}$ but rarely in AFS1$_{bottom}$, then it has a high weight. Note that the style of \autoref{E4} is '$vector = vector - vector$'. This concise formula is the heart of NFFS. What the 'normalized frequencies' in the title of the paper refers to is \autoref{E4}.

From the above step, we know that the items in the vector \overrightarrow{F_{top}} and vector \overrightarrow{F_{bottom}} are the frequencies of the features. It makes sense to compare a frequency in \overrightarrow{F_{top}} with another frequency in \overrightarrow{F_{top}}. But it is meaningless to compare a frequency in \overrightarrow{F_{top}} with a frequency in \overrightarrow{F_{bottom}}. This is why normalization is needed.

\textbf{3. Generate AFS2 by \emph{sorting}}: When the WV2 is obtained, it is also the time to generate feature subsets, the generation process is much simpler than those in phase I of NFFS.

The first feature subset generated is the feature with the highest weight in WV2, the second feature subset generated is the two features with highest weight in WV2, the third feature subset generated is the three features with the highest weight in WV2, and so on. In total, O (O $<$ number of features) feature subsets is generated.These feature subsets constitute AFS2.

\textbf{4. Evaluate feature subsets and get the result}: Each feature subset in AFS2 is provided to the classifier for evaluation, and the feature subset with the highest fitness value would be selected as the output of NFFS. \autoref{Phase II of NFFS} show an overall procedure for Phase II of NFFS. NFFS only needs to evaluate ($L + O$) feature subsets to obtain result, which is a great advantage in speed compared to the heuristic algorithms, and the result is excellent.

\begin{algorithm}
\caption{Phase II of NFFS}\label{Phase II of NFFS}
\KwIn{$M, N, O$}
\KwResult{Best feature subset.}
Sort feature subsets in AFS1 by their fitness values.\\
Get AFS1$_{top}$ and AFS1$_{bottom}$ from AFS1.\\
Get $\overrightarrow{F_{top}}$ from AFS1$_{top}$.\\
Get $\overrightarrow{F_{bottom}}$ from AFS1$_{bottom}$.\\
Use \autoref{E4} WV2.\\
%Generate AFS2.
\While{$O>0$}{AFS2$_O$ = The $O$ feature subsets with the highest weight in WV2.\\ \tcp*[h]{AFS2$_O$  represents the $O$-th feature subset in AFS2.}\\$O$=$O$-1}
Evaluate the feature subsets inside AFS2.\\
Sort feature subsets in AFS2 by their fitness values.\\

\Return Feature subset that get the best fitness value in AFS2.
\end{algorithm}

\section{Experiments and results}
\label{S5}
In his section, we introduced the dataset used for experiments. Data preprocessing is discussed. We described the evaluation indicators and classifier used in this paper, fitness function is also be illustrated. In \autoref{S5.4}, we implemented the proposed NFFS. In \autoref{S5.5}, we described the results and compared the proposed method with several state-of-the-art feature selection methods.
% In \autoref{S5.5}, we use experiments to illustrate why we generate feature subsets differently in phase I and II of NFFS.

\subsection{Dataset}
\label{S5.1}
NSL-KDD dataset \citep{R27} and UNSW-NB15 \citep{R27} dataset are used to evaluate the proposed feature selection method. These two dataset are authoritative real-world dataset for intrusion detection domain. Also both datasets have been provided with ready-made training dataset and testing dataset. It is not necessary to prepare the training dataset and test dataset by sampling from the unsegmented datase.

NSL-KDD dataset uses 41 features to represent a record, and each record is either an normal or attack. This dataset is a refined version of KDD Cup 99 dataset \citep{R261} and adds a item to represent the difficulty of classifying correctly.  UNSW-NB15 dataset consists of 42 features and a multiclass label and a binary label, here we only use the binary label. There are no duplicate records in these two dataset.

\begin{table}
\centering
\caption{Type of features in NSL-KDD dataset.}
\label{TK}
\begin{tabular}{ll}
\toprule
\multicolumn{1}{c} {Types of features} & \multicolumn{1}{c}{Features}                                                                                                                                                                                                                                                                                      \\ \midrule
Binary                     & [ f7,  f12,  f14, f15, f21, f22 ]                                                                                                                                                                                                                                                                            \\ 
Categorical                & [ f2, f3, f4 ]                                                                                                                                                                                                                                                      \\ 
Numeric                    & \multicolumn{1}{p{0.8\columnwidth}}{[ f1, f5, f6, f8, f9, f10, f11, f13, f16, f17, f18, f19, f20, f23, f24, f25, f26, f27, f28, f29, f30, f31, f32, f33, f34, f35, f36, f37, f38, f39, f40, f41 ]}                                                                                                                                                                                                                                                                                     \\ \bottomrule
\end{tabular}
\end{table}

\begin{table}
\centering
\caption{Type of features in UNSW-NB15 dataset.}
\label{TU}
\begin{tabular}{ll}
\toprule
\multicolumn{1}{c} {Types of features} & \multicolumn{1}{c}{Features}                                                                                                                                                                                                                                                                                      \\ \midrule
Binary                     & [ f37, f42 ]                                                                                                                                                                                                                                                                            \\ 
Categorical                & [ f2, f3, f4 ]                                                                                                                                                                                                                                                      \\ 
Numeric                    & \multicolumn{1}{p{0.8\columnwidth}}{[ f1, f5, f6, f7, f8, f9, f10, f11, f12, f13, f14, f15, f16, f17, f18, f19, f20,  f21, f22, f23, f24, f25, f26, f27, f28, f29, f30, f31, f32, f33, f34, f35, f36, f38, f39, f40, f41 ]}                                                                                                                                                                                                                                                                                     \\ \bottomrule
\end{tabular}
\end{table}

\begin{table}[]
\centering
\caption{Summary quantitative information of NSL-KDD datatset and UNSW-15 dataset.}
\label{TS}
%\resizebox{0.8\textwidth}{!}{%
\begin{tabular}{cccc}
\toprule
Datasat                           & Partition        & Positive (Ratio) & Negative (Ratio) \\ \midrule
\multirow{2}{*}{NSL-KDD} & Training dataset & 58630 (46.5\%)  & 67343 (53.5\%)  \\
                                  & Testing dataset  & 12833 (56.9\%)  & 9711 (43.1\%)   \\
\multirow{2}{*}{UNSW-15} & Training dataset & 119341 (68.1\%) & 56000 (31.9\%)  \\
                                  & Testing dataset  & 45332 (55.1\%)  & 37000 (44.9\%) \\  \bottomrule
\end{tabular}%
%}
\end{table}

The features of NSL-KDD dataset are presented in \autoref{TK}  in a manner that is more friendly to machine learning \citep{R31}. Where 'f$n$' represents the column $n$ in the dataset file. \autoref{TU} show the features of UNSW-NB15 dataset in the same style. As shown in \autoref{TK} and \autoref{TU}, both dataset contains three categorical features. \autoref{TS} shows the distribution for NSL-KDD and UNSW-NB15. As the table shows that both dataset are relatively balanced. 

\subsection{Data preprocessing}
\label{S5.2}
The first step in the experiments is to process the nonnumerical marks in dataset. Preprocessing consists of two items: convert categorical features into numeric features by one-hot encoding; convert symbolic labels into binary labels. In the experiments, we used '0' denotes normal class while '1' denotes attack class in spite of the specific type of attack. After one-hot encoding processed, 41 features inside NSL-KDD have been expanded to 122.

\subsection{Classifier related}
\label{S5.3}
\subsubsection{Classifier}
In this paper, random forest \citep {R32} with PCA \citep{R33} is used as the classifier, and PCA is regarded as a part of the classifier rather than an independent processing step \citep{R331}. All feature subsets need to be evaluated by this classifier. The PCA plays an important role as part of the classifier, the data will be processed by PCA before being fed into the classifier. 

PCA can project data into an orthogonal feature space, which can remove redundant linear relationships from the original feature space. PCA makes the classifier robuster and at the same time further decreases the computational cost.

Since PCA is a scale sensitive method, it needs the help of standardization. It is important to note that CART-based random forest does not need standardization, standardization is only for PCA here. This classifier is also used to evaluate the feature selection methods from state-of-the-art related works.

\subsubsection{Evaluation indicators}
Different learning tasks in machine learning require different indicators, the evaluation indicators in this paper adopt generic nomenclature, rather than exclusive to a specific task. Accuracy, recall, precision, F-score and AUC are used to evaluate feature subsets in this paper. All indicators used can be calculated from the confusion matrix \citep{R34}, the confusion matrix is shown in \autoref{T3}, where the positive means attack class, while the negative means normal class. The four parameters in table represent the number of records that match a certain condition:

\begin{table}
\centering
\caption{Binary confusion matrix.}
\label{T3}
\begin{tabular}{lcc}
\toprule
           & Predicted normal & Predicted attack \\ \midrule
Actual normal & TN          & FP          \\ 
Actual attack & FN          & TP           \\ \bottomrule
\end{tabular}
\end{table}

\textbf{TP (True Positive)}: Attack and predicted to be attack.

\textbf{TN (True Negative)}: Normal and predicted to be normal.

\textbf{FP (False Positive)}: Normal and predicted to be attack.

\textbf{FN (False Negative)}: Attack and predicted to be attack.

The evaluation indicators used in experiments are described as follows:

\textbf{1. Accuracy}: Measure how many records are correctly classified as in \autoref{E5}.
\begin{equation}
\label{E5}
\large
accurary=\frac{TP+TN}{TP+TN+FP+FN}
\end{equation}

\textbf{2. Recall}: Measures how many attacks could be discovered as in \autoref{E6}.
\begin{equation}
\label{E6}
\large
recall=\frac{TP}{TP+FN}
\end{equation}

\textbf{3. Precision}: Measures how many attacks classified as attack are really attack as in \autoref{E7}.
\begin{equation}
\label{E7}
\large
precision=\frac{TP}{TP+FP}
\end{equation}

\textbf{4. F-score}: A weighted average of the precision and recall as in \autoref{E8}.
\begin{equation}
\label{E8}
\large
F-score=2\times\frac{\left(precision*recall\right)}{\left(precision+recall\right)}
\end{equation}

\textbf{5. AUC(Area Under the Receiver Operating Characteristic)}: Simply put it monitors the potential of the model \citep{R35}.

Recall and precision are two of the mutually exclusive evaluation indicators. A classifier can only achieve a high F-score if it obtains high values on both recall and precision. All evaluation indicators except AUC are in the range [0,1]. The maximum value of AUC is 1, while the minimum value is 0.5. If a classifier makes a random decision, then its AUC will be 0.5.

\subsubsection{Fitness function}
In the search for the optimal feature subset, we use only F-score to constitute fitness function without considering the number of features in feature subset, which is based on two considerations: the F-score is a comprehensive indicator; and NFFS is not a wrapper method, there is no comparison of independent feature subsets. Note that fitness function is used to search for the optimal feature subset, while evaluation indicators are used to evaluate the final result.

\subsection{Perform feature selection}
\label{S5.4}
Scikit-learn is a Python module for machine learning. In this paper, experiments were completed by Scikit-learn \citep{R39}. The quality of a feature subset needs to be judged by its fitness value, but the fitness value from the classifier is highly dependent on the parameters used to train the classifier. To solve this problem, we adopt the strategy of testing a feature subset with multiple groups of parameters, and the average value is used as the fitness value of the feature subset. 

\subsubsection{Phase I of NFFS}
\label{S5.4.1}

\begin{figure}
	\centering
	\includegraphics[width=0.4\textwidth]{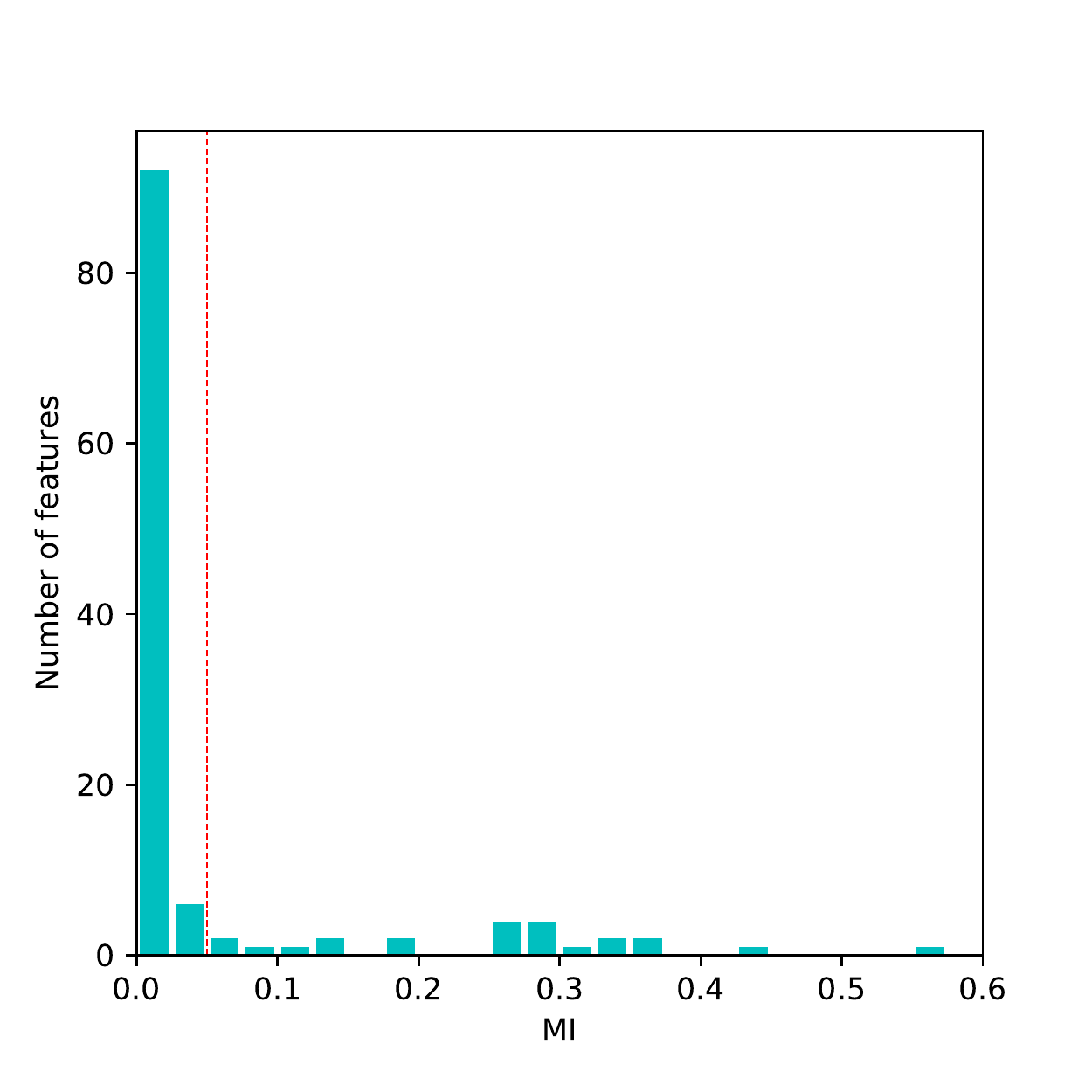}
	\caption{Distribution for MI values of features in NSL-KDD dataset.}
	\label{F4}
\end{figure}

\autoref{F4} shows the MI values of features in NSL-KDD dataset, from which it can be seen that the majority of features hold tiny MI values. The threshold taken in NSL-KDD dataset is 0.05, and it is indicated by a red vertical line in \autoref{F4}. As shown in figure, a large number of features would hold a weight of 0.5 after calculation by \autoref{E2}. However, the final optimal feature subset show that majority of features are eliminated as well.

\begin{figure}
	\centering
	\includegraphics[width=0.4\textwidth]{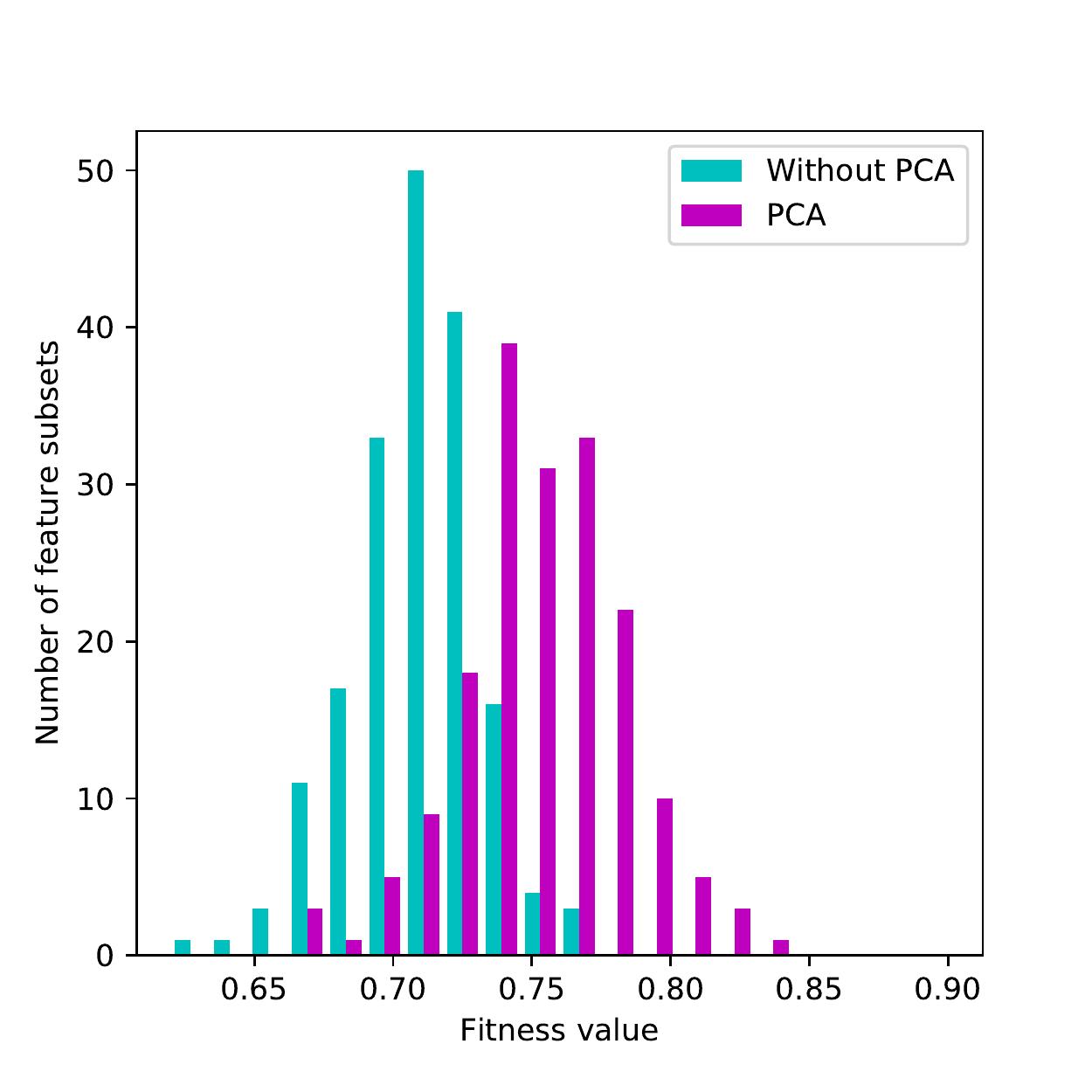}
	\caption{Distribution for fitness values of feature subsets in AFS1.}
	\label{F5}
\end{figure}

After obtaining WV1 via \autoref{E2}, \autoref{E3} was used to generate 180 feature subsets, which constituted AFS1. We wanted to get a glimpse of the power of PCA. \autoref{F5} shows the performance of feature subsets in AFS1. As can be observed in the figure, the classifier with PCA shifts the bar significantly to the right, which means fitness values of feature subsets is significantly improved by PCA. The reason is that PCA removes linearly correlated redundant information from the input data, which allows a robuster classifier. Note that, this is just to a test of the significance of PCA, elsewhere PCA always accompanies classifier around.
\subsubsection{Phase II of NFFS}
\label{S5.4.2}
In experiments, 45 feature subsets with higher fitness values from AFS1 were selected to constitute AFS1$_{top}$ and 45 features with lower F-score were selected to constitute AFS1$_{bottom}$. After counting frequencies, WV2 can be calculated by \autoref{E4}, which were used to generate 70 feature subsets to constitute AFS2.

\begin{figure}
	\centering
	\includegraphics[width=0.4\textwidth]{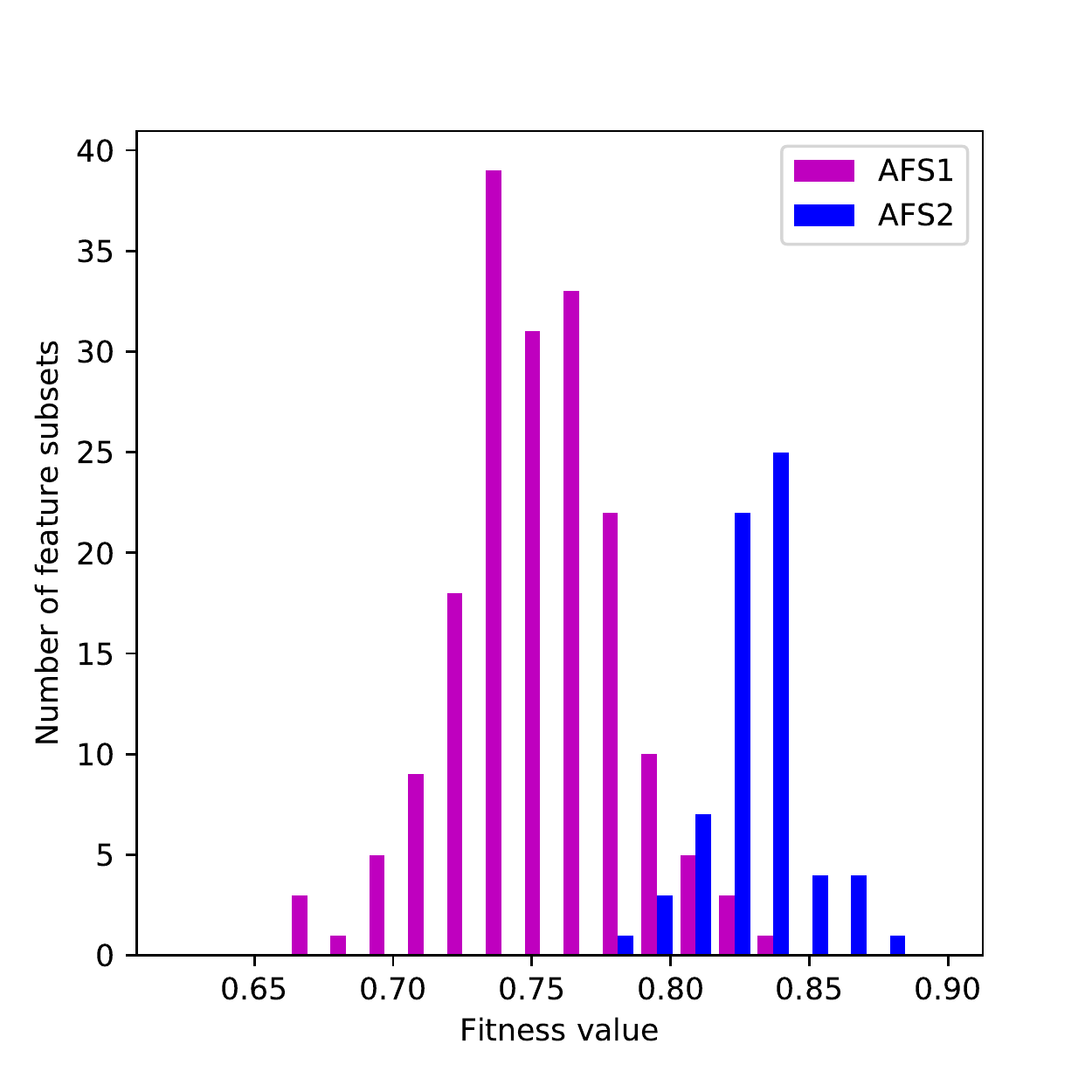}
	\caption{Distribution for fitness values of feature subsets in AFS1 and AFS2.}
	\label{F6}
\end{figure}

   \autoref{F6} shows the performance of feature subsets in AFS1 and AFS2. As figure shown, feature subsets in AFS2 hold distinct advantages, which suggests that phase II of NFFS plays a significant role. It can also be seen that the feature subsets in AFS2 performed more stably. The phase II of NFFS is based on the phase I of NFFS but is far superior to it.

\subsection{Result}
\label{S5.5}
\begin{table}
\centering
\caption{Results of several feature selection methods applied to the NSL-KDD dataset.}
\label{T4}
\begin{tabular}{llcp{0.36\columnwidth}}
\toprule
\multicolumn{1}{c}{Reference} & \multicolumn{1}{c}{Method}        & NF & \multicolumn{1}{c}{Selected feature subset}                                                                                                                                                                                                                                                                                                          \\ \midrule 
          \citep{enache2015anomaly}         & BAT                       & 89(18)                          &[ f1, f2, f3, f8, f9, f13, f14, f18, f19, f20,  f26, f28, f32, f33, f34, f38, f39, f40 ]                                                                                                                                                                                                                                                   \\ 
            \citep{ambusaidi2016building}       & LSSVM                     & 97(18)                          &[ f3, f4, f5, f6, f12, f23, f25, f26, f28, f29,  f30, f33, f34, f35, f36, f37, f38, f39 ]                                                                                                                                                                                                                                                  \\ 
            \citep{moustafa2015a}       & Hybrid Association Rules  & 13(11)                          &[ f2, f5, f6, f7, f12, f16, f23, f28, f31, f36,  f37 ]                                                                                                                                                                                                                                                                              \\ 
       \citep{aljawarneh2017anomaly-based}      & IG                        & 87(8)                           & [ f3, f4, f5, f6, f29, f30, f33, f34 ]                                                                                                                                                                                                                                                                                                                                      \\
            \citep{tama2019tse-ids:}       & PSO                       & 118(37)                          &[ f2, f3, f4, f5, f6, f7, f8, f9, f10, f11, f12,  f13, f14, f15, f17, f18, f20, f21, f22, f23,  f24, f25, f26, f27, f28, f29, f31, f32, f33,  f34, f35, f36, f37, f38, f39, f40, f41 ]                                                                                                                                                                    \\ 
       \citep{R26}            & Sigmoid\_PIO              & 98(18)                          &[ f1, f3, f4, f5, f6, f8, f10, f11, f12, f13, f14,  f15, f17, f18, f27, f32, f36, f39, f41 ]                                                                                                                                                                                                                                                \\ 
         \citep{R26}          & Cosine\_PIO               & 7(5)                           & [ f2, f6, f10, f22, f27 ]                                                                                                                                                                                                                                                                                                                                                \\           Proposed method        & NFFS                      & 34(11)                      &[ f2\_icmp, f3\_IRC, f3\_aol, f3\_auth, f3\_csnet\_ns, f3\_ctf, f3\_daytime, f3\_discard, f3\_ecr\_i, f3\_http\_8001, f3\_imap4, f3\_login, f3\_name, f3\_netbios\_ssn, f3\_pop\_2, f3\_pop\_3, f3\_rje, f3\_supdup, f3\_telnet, f3\_urh\_i, f4\_OTH, f4\_RSTO, f4\_RSTOS0, f4\_RSTR, f4\_S1, f4\_SF, f6, f7, f9, f10, f21, f30, f40, f41 ]\\ \bottomrule
\end{tabular}
\end{table}

The related works used to compared with NFFS and the features they selected from NSL-KDD are summarized in \autoref{T4}. Where the number outside the parentheses in the third column (NF) denote the number of encoded features (encoded by one-hoting encoding), while the number inside the parentheses denote the number of original features. It can be noticed that the feature subset selected by NFFS is reported with the encoded format, that is because NFFS is selecting features directly from the encoded feature space. Regarding the format of the features selected by NFFS, for example, f$2$\_tcp indicates the category ‘tcp’ in the 2-th feature (communication protocol).

In order to fairly compare the quality of the feature subsets found by each feature selection method, we need to train a separate classifier for each feature subset. For each feature subset reported in \autoref{T4},  we searched for the best parameters to build a classifier for it. During the search for the optimal parameters, the explanation ratio of PCA was always set to 0.93. Thirty different random seeds was used to perform 30 runs to obtain means and standard deviations to compose the results of the method. The 30 random seeds used are integers from 7 to 36, and the random seeds start from 7 simply because we believe that 7 represents luck. The fixed random seed allows the experimental results to be accurately reproduced. %The number and depth of trees adopted in random forest are given in \autoref{T6}, other unspecified parameters adopted the default in Scikit-learn.

Each feature subset in \autoref{T4} was fed to the customized classifier for evaluation, and the evaluation results are presented in \autoref{T5}. Based on the results shown in \autoref{T5}, NFFS obtain the best score in terms of accuracy, precision, recall, F-score and AUC.

\begin{table}[]
\centering
\caption{The performances of the feature subsets in \autoref{T4}.}
\label{T5}
\begin{tabular}{lccccc}
\toprule
Methods          & Precision±std & Recall±std & Accuracy±std & F-score±std & AUC±std \\ \midrule
BAT                   & 0.962±0.011          & 0.704±0.029       & 0.815±0.016         & 0.812±0.019        & 0.838±0.014    \\ 
LSSVM                     & 0.904±0.002          & 0.842±0.019       & 0.859±0.010         & 0.871±0.011        & 0.921±0.007    \\ 
Hybrid Association Rules& 0.956±0.009          & 0.636±0.020       & 0.776±0.011         & 0.763±0.015        & 0.802±0.012    \\ 
IG                & 0.898±0.002          & 0.781±0.032       & 0.825±0.017         & 0.835±0.018     & 0.927±0.005   \\ 
PSO                     & 0.948±0.021          & 0.687±0.035       & 0.800±0.021         & 0.796±0.025        & 0.819±0.020   \\ 
Sigmoid\_PIO             & 0.921±0.001         & 0.702±0.011       & 0.796±0.006         & 0.797±0.007       & 0.880±0.022   \\ 
Cosine\_PIO             & 0.926±0.003          & 0.821±0.040       & 0.861±0.022         & 0.870±0.023       & 0.910±0.009 \\ 
NFFS                      & 0.963±0.005          & 0.852±0.028       & 0.897±0.015         & 0.904±0.016        & 0.938±0.001    \\ \bottomrule
\end{tabular}
\end{table}

\begin{figure}
	\centering
	\includegraphics[width=0.8\textwidth]{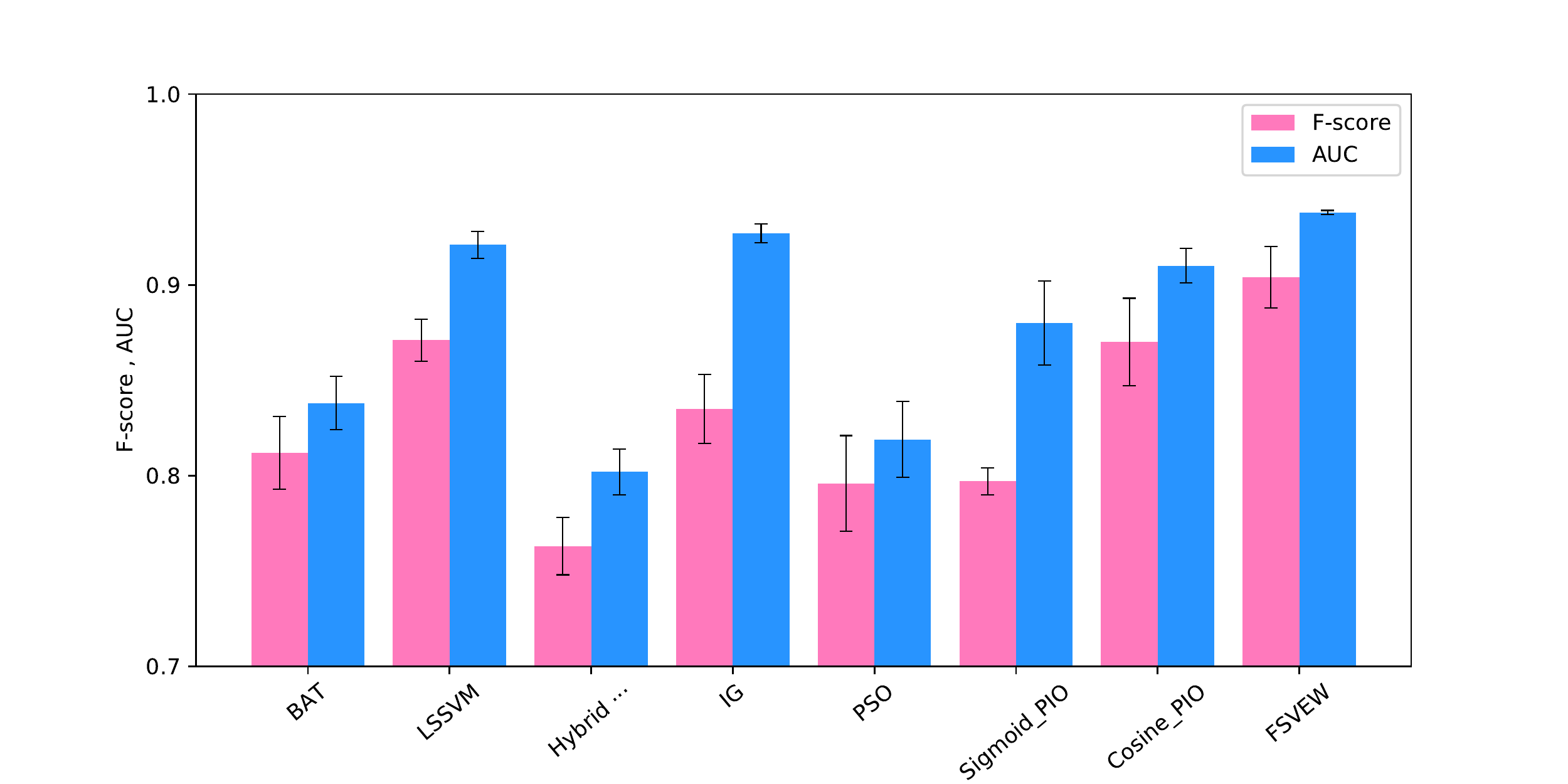}
	\caption{Performances of feature selection methods in \autoref{T4}.}
	\label{F7}
\end{figure} 
F-score and AUC are relatively more comprehensive and pertinent indicators in machine learning, and \autoref{F7} visualizes these two indicators from \autoref{T5}. Where each bar represent mean and standard deviation of each method. As shown in \autoref{F7}, NFFS achieves the highest F-score and AUC. For F-score, NFFS gains an absolute victory without any doubt. For AUC, NFFS not only achieved the highest score, but also hold the smallest standard.

\begin{table}
\centering
\caption{Results of several feature selection methods applied to the UNSW-NB15 dataset.}
\label{Results of UNSW}
\begin{tabular}{llcp{0.36\columnwidth}}
\toprule
\multicolumn{1}{c}{Reference} & \multicolumn{1}{c}{Method}        & NF & \multicolumn{1}{c}{Selected feature subset}                                                                                                                                                                                                                                                                                                          \\ \midrule 
            \citep{moustafa2015a}       & Hybrid Association Rules  & 21(11)                          &[ f4, f10, f11, f18, f20, f23, f25, f32, f35, f40, f41 ]                                                                                                                                                                                                                                                                              \\ 
            \citep{tama2019tse-ids:}       & PSO                       & 41(19)                          &[ f3, f4, f7, f8, f10, f11, f16, f20, f22, f24, f26, f28, f30, f32, f34, f35, f36, f41, f42 ] \\           
            \citep{tama2019tse-ids:}       & Rule-Based                       & 157(13)                          &[ f2, f3, f7, f8, f10, f15, f17, f31, f33, f34, f35, f36, f41 ]  \\           
Proposed method        & NFFS                      & 38(14)                      &[ f7, f10, f11, f14, f19, f27, f28, f29, f34, f35, f36, f2\_argus, f2\_bbn-rcc, f2\_br-sat-mon, f2\_cbt, f2\_crudp, f2\_dcn, f2\_ddp, f2\_eigrp, f2\_hmp, f2\_ipcv, f2\_leaf-2, f2\_netblt, f2\_ospf, f2\_ptp, f2\_scps, f2\_snp, f2\_st2, f2\_vines, f3\_pop3, f3\_ssl, f4\_CLO, f4\_CON, f4\_no ]\\ \bottomrule
\end{tabular}
\end{table}

UNSW-NB-15 is another dataset used to evaluate the NFFS.  \autoref{Results of UNSW} reports the selected feature subsets from UNSW-NB15 dataset. The format of \autoref{Results of UNSW} is the same as that of \autoref{T4}. \autoref{POU} show the results of realted works and NFFS. The format of \autoref{POU} is the same as that of \autoref{T5}, the data therein are also derived from 30 runs. As the table shows, NFFS get the best result against other method in term of the five indicators. 

\begin{table}[]
\centering
\caption{The performances of the feature subsets in \autoref{Results of UNSW}.}
\label{POU}
\begin{tabular}{lccccc}
\toprule
Methods          & Precision±std & Recall±std & Accuracy±std & F-score±std & AUC±std \\ \midrule
Hybrid Association Rules   & 0.762±0.001          & 0.968±0.001       & 0.816±0.001         & 0.853±0.001        & 0.937±0.001    \\ 
PSO                    & 0.768±0.003          & 0.983±0.004       & 0.827±0.003         & 0.862±0.002        & 0.960±0.002   \\ 
Rule-Based     & 0.797±0.002          & 0.973±0.003       & 0.849±0.002         & 0.876±0.002       & 0.962±0.002   \\ 
NFFS                      &  0.818±0.004          & 0.985±0.001       & 0.871±0.003         &0.893±0.002        & 0.977±0.002    \\ \bottomrule
\end{tabular}
\end{table}

\section{Conclusion}
\label{S6}
In this work, a concise method for feature selection is proposed to overcome the shortcomings of metaheuristic algorithms. The proposed method is divided into two phases to implement feature selection. The proposed method is based on the following ideas: (Phase I) If the filter method considers a feature of higher importance, then the feature is selected with higher probability in the generation of feature subsets; (Phase II) If a feature is often contained by the feature subsets that perform well, while rarely being contained by the feature subsets that perform poorly, it means that the feature is beneficial to the classifier, the opposite means that the feature is harmful to the classifier.

In order to provide sufficient feature subsets with diverse performances for phase II. We generated the feature subsets by probability in phase I, which allowed the generated feature subsets better than the randomly generated ones. We used experiments to illustrate the reasons why the two phases used different strategies to generate feature subsets. The experimental results indicate that the proposed method outperformed several methods from state-of-the-art related works in terms of precision, recall, accuracy, F-score and AUC.

Future researches can investigate the use of clustering to optimize the efficiency of encoding and the application of the proposed method to semi-supervised learning.

%\section*{CRediT authorship contribution statement}
%\textbf {Tan Song}: Conceptualization, Methodology, Software, Writing - Original Draft, Visualization. \textbf {Cui Jiyuan}: Software, Validation, Data Curation. \textbf {Mao hongyun}: Software, Formal analysis, Supervision. \textbf {Xie Jianhua}: Validation, Data Curation. \textbf {Tang Mingwei}: Writing - Review \& Editing, Supervision, Project administration.
%
%
%\section*{Declaration of competing interest}
%The authors declare that they have no known competing financial interests or personal relationships that could have appeared to influence the work reported in this paper.

\bibliographystyle{elsarticle-harv} 
\bibliography{ref.bib} 
\end{document}